%% file: anonymous-submission-latex-2026.tex
\documentclass[letterpaper]{article} % DO NOT CHANGE THIS
\usepackage{aaai2026}  % DO NOT CHANGE THIS
\usepackage{times}  % DO NOT CHANGE THIS
\usepackage{helvet}  % DO NOT CHANGE THIS
\usepackage{courier}  % DO NOT CHANGE THIS
\usepackage[hyphens]{url}  % DO NOT CHANGE THIS
\usepackage{graphicx} % DO NOT CHANGE THIS
\usepackage{natbib}  % DO NOT CHANGE THIS AND DO NOT ADD ANY OPTIONS TO IT
\usepackage{caption} % DO NOT CHANGE THIS AND DO NOT ADD ANY OPTIONS TO IT
\usepackage{algorithm}
\usepackage{algorithmic}
\usepackage{booktabs}
\usepackage{makecell}
\usepackage{url}
\usepackage{framed}
\usepackage{newfloat}
\usepackage{listings}
\DeclareCaptionStyle{ruled}{labelfont=normalfont,labelsep=colon,strut=off} % DO NOT CHANGE THIS
\floatstyle{ruled}
\newfloat{listing}{tb}{lst}{}
\floatname{listing}{Listing}
\nocopyright
\title{PRGB Benchmark: A Robust Placeholder-Assisted Algorithm for\\ Benchmarking Retrieval-Augmented Generation}
\author{
    Zhehao Tan\equalcontrib, Yihan Jiao\equalcontrib, Dan Yang, Lei Liu\\
    Jie Feng, Duolin Sun, Yue Shen, Jian Wang, Peng Wei, Jinjie Gu
}
\affiliations{
    Ant Group\\
    \{tanzhehao.tzh, jiaoyihan.yh, luoyin.yd\}@antgroup.com 
}
\usepackage{bibentry}
\begin{document}

\maketitle

\begin{abstract}
Retrieval-Augmented Generation (RAG) enhances large language models (LLMs) by integrating external knowledge, where the LLM's ability to generate responses based on the combination of a given query and retrieved documents is crucial. However, most benchmarks focus on overall RAG system performance, rarely assessing LLM-specific capabilities. Current benchmarks emphasize broad aspects such as noise robustness, but lack a systematic and granular evaluation framework on document utilization. To this end, we introduce Placeholder-Rag-Benchmark, a multi-level fine-grained benchmark, emphasizing the following progressive dimensions: (1) \textit{\textbf{multi-level filtering abilities}}, %evaluating the LLM's capacity to filter noisy documents progressively; 
(2) \textit{\textbf{combination abilities}}, %examining the LLM's proficiency in synthesizing information between multiple entities or attributes within single or multiple documents; 
and (3) \textit{\textbf{reference reasoning}}. 
%including contrastive and inductive reasoning based on both retrieved documents and the model’s parametric knowledge. 
To provide a more nuanced understanding of LLMs' roles in RAG systems, we formulate an innovative \textit{\textbf{Placeholder}}-based approach to decouple the contributions of the LLM's parametric knowledge and the external knowledge. Experiments demonstrate the limitations of representative LLMs in the RAG system's generation capabilities, particularly in error resilience and context faithfulness. Our benchmark provides a reproducible framework for developing more reliable and efficient RAG systems. Our code is available in \url{https://github.com/Alipay-Med/PRGB}.
\end{abstract}

\input{sec/1_intro}

\input{sec/3_benchmark}
\input{sec/4_exp}
\input{sec/5_conclusion}

\newpage
\bibliography{aaai2026}

\newpage
\appendix
\section{Appendix: Dataset Analysis}
\label{sec:data_des}
The data details for different subtasks are presented in Table \ref{tab:details of subtasks-en} and Table \ref{tab:details of subtasks-zh} . These tables provide comprehensive information about each task, including the total data volume, the average number of golden documents, the average number of noisy documents, and the average count of candidate placeholders.

\begin{table}[H]
\centering
{
\begin{tabular}{lcccccc}
\toprule
\textbf{Task} & Quantity& Avg.golden doc &Avg.weak noise & Avg.moderate noise & Avg.hard noise &Avg.placeholders\\
\midrule
\texttt{Filtering}  & 816 & 1.00 & 5.00 & 4.84 & 0.86 &4.00\\
\texttt{Combination} & 2042 & 2.42 & 5.00 & 3.73 & 0.68&4.00\\
\texttt{Reasoning} & 1029 & 1.00 & 7.88 & 4.17 & 0.00&5.26\\
\bottomrule
\end{tabular}
}
\caption{Data details of subtasks in the English Dataset}
\label{tab:details of subtasks-en}
\end{table}

\begin{table}[H]
\centering
{
\begin{tabular}{lcccccc}
\toprule
\textbf{Task} & Quantity& Avg.golden doc &Avg.weak noise & Avg.moderate noise & Avg.hard noise &Avg.placeholders\\
\midrule
\texttt{Filtering}  & 800 & 1.00 & 5.00 & 4.80 & 0.84 &4.00\\
\texttt{Composition} & 1495 & 2.53 & 5.00 & 3.14& 0.608&4.00\\
\texttt{Reasoning} & 1092 & 1.81 & 8.60 & 4.61 & 0.25&5.92\\
\bottomrule
\end{tabular}
}
\caption{Data details of subtasks in the Chinese Dataset}
\label{tab:details of subtasks-zh}
\end{table}
\end{document}

%% file: sec/1_intro.tex
\section{Introduction}

\begin{figure*}
\centering
\includegraphics[height=10cm, width=2\columnwidth]{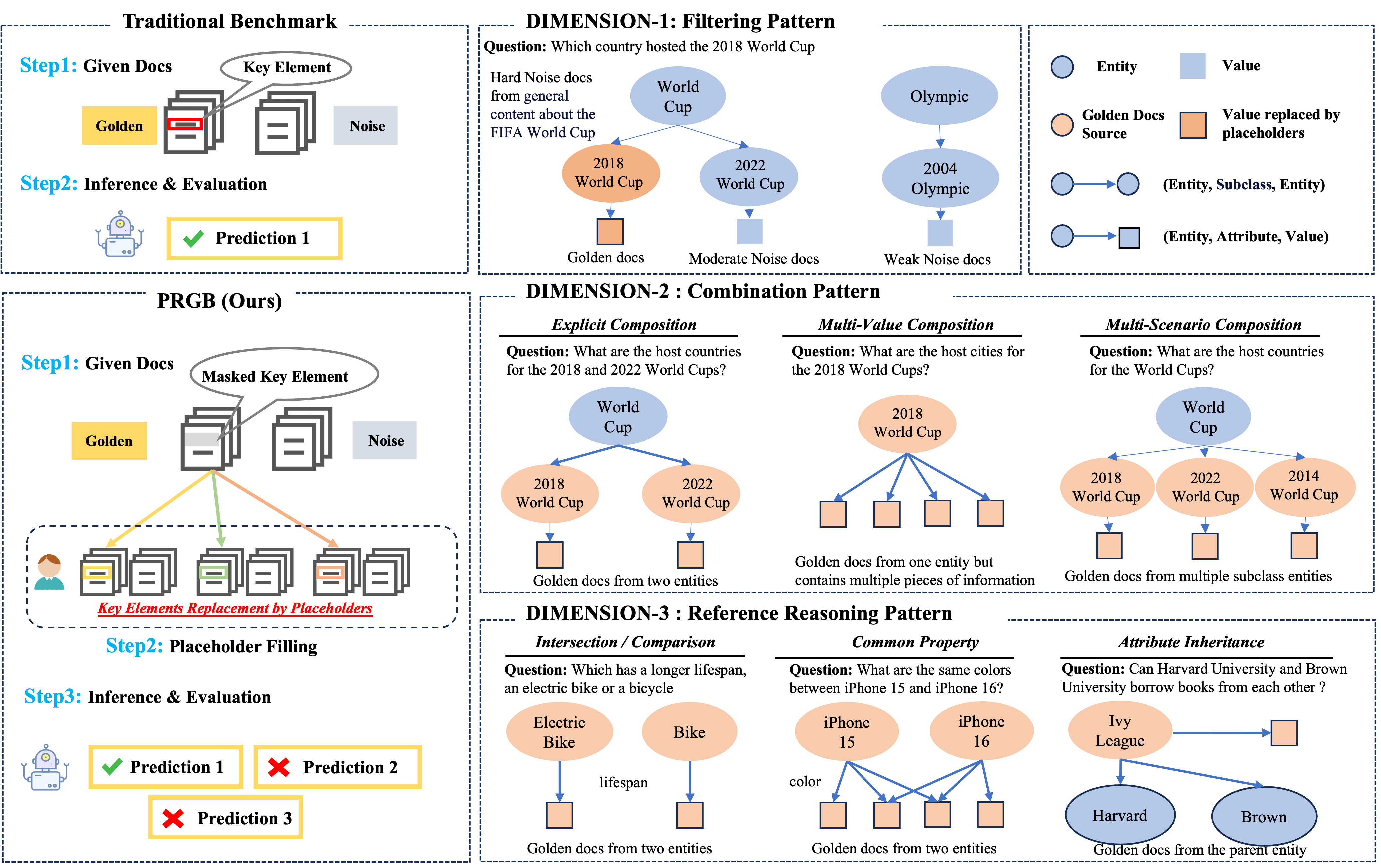}
    \caption{Visualization for Evaluation Dimensions of Placeholder-RAG-Eval. Starting from triplet-based metadata, three finer-grained evaluation tasks are formulated, including filtering, composition, and reasoning.}
    \label{fig:pls}
\end{figure*}

Retrieval-Augmented Generation (RAG) has emerged as a transformative paradigm enabling large language models (LLMs) to dynamically integrate external knowledge, where it is crucial for LLMs within RAG systems to generate more reliable responses based on the combination of a given query and retrieved documents.

Recently, diverse efforts have been conducted to evaluate RAG systems' performance across methodologies and metrics. For example, RAG \cite{lewis2020retrieval}, REALM \cite{lewis2020retrieval}, and RETRO \cite{borgeaud2022improving} assess retrieval quality via exact match or F1 scores over retrieved passages and generation accuracy through QA-based metrics. However, existing benchmarks predominantly evaluate the overall performance of RAG systems while lacking in-depth exploration of LLMs' generation capabilities based on the retrieved documents.

Upon foundational theories in information retrieval evaluation \cite{voorhees1998variations}, which emphasize the dual importance of precision (relevance) and recall (completeness), we extend these principles to the RAG context by explicitly modeling their interplay with generative capabilities. This aligns with recent findings that retrieval quality alone cannot guarantee generation accuracy \cite{min2020ambigqa}. Therefore, it is very necessary to \textbf{evaluate the capabilities of generative models in RAG systems}. HIRAG\cite{hirag} trained the generation models in RAG systems, but it lacks more granular aspects about RAG-specific tasks. To investigate LLMs' capabilities to leverage RAG context, RAG-Bench \cite{RGB} systematically combined pre-defined RAG metrics and newly introduced LLM-specific metrics (\textit{e.g.}, fluency, coherence, and factual consistency) to better describe the overall RAG system
performance. RGB \cite{rgb_xy} tried to define basic abilities of RAG-based LLMs from the viewpoints of noise robustness and document integration. However, the dimensions of these benchmarks are relatively coarse, which is \textbf{insufficient to evaluate the capability of LLMs to effectively leverage the retrieved corpus within RAG systems}. The resulting capabilities that LLMs should possess in RAG environments are as follows: \textbf{(a) Multi-level Filtering} evaluates whether LLMs can accurately identify the relevant information from retrieved documents, filtering varying-degree noise. \textbf{(b) Combination} evaluates whether LLMs can recognize and retain all relevant data necessary for generating a complete and accurate response. \textbf{(c) Reference Reasoning} evaluates whether LLMs can perform multi-hop reasoning based on the information provided in retrieved documents and answer questions that go beyond direct answers on content retrieval (the above two abilities).
%Precision in Information Retrieval， Comprehensiveness in Information Retrieval， Information Processing

% the model’s ability to utilize the information effectively. When the information provided in documents is insufficient to directly answer a question, the model must perform some level of inference based on the available data. 

 % essentially measuring its capability to filter out noise. The goal is to ensure that the model can accurately pinpoint the necessary data amidst potentially distracting information.
 % ability to be thorough, emphasizing its capacity to avoid mistakenly filtering out pertinent information. It measures how well the model can 

% Basically, the distinction between precision and comprehensiveness in retrieval echoes the classical tension between specificity and coverage in knowledge base construction \cite{suchanek2007yago}, where overly strict filtering risks losing critical facts while excessive inclusion introduces noise.

However, when evaluating whether LLMs could genuinely ground responses in external retrieval contexts, a critical challenge lies in \textbf{disentangling the contribution of LLMs' parametric knowledge}, since conflating these knowledge sources would obscure whether LLMs default to applying pre-trained parametric biases. For instance, when a model correctly answers a factual question using retrieved documents, it remains unclear whether the accuracy stems from genuine synthesis of external information or implicit recall of pre-trained knowledge.
Recent solution POPQA \cite{pqa} used less popular factual data for evaluation, because such data is more likely to be learned by LLMs. However, over time, newer LLMs are likely to learn from an increasing volume of online corpora, which means that information that was relatively less popular at the time may eventually be memorized by these models after a certain period. When evaluating datasets with RAG like TriviaQA \cite{tqa} and RGB\cite{rgb_xy}, the majority of methods adopted a specific instruction template ``If the information in the document does not contain the answer, you will generate I cannot answer the question because of insufficient information in the documents''. However, instruction would influence the RAG evaluation, because LLMs with stronger instruction following ability are more likely to refuse to answer with ``I cannot answer the question".

% However, based on our experiments with TQA\cite{tqa} depicted in ~\ref{fig:comparison}, more powerful models with stronger adherence to commands are more likely to respond with "I cannot answer the question," making it challenging to intuitively assess the true retrieval-augmented generation (RAG) capability through prompting alone.

% approach by instructing

% Finally, in order to better focus on testing the model's ability to utilize information, however, this approach has certain requirements for timeliness and necessitates continuous updates, as over time, non-parametric knowledge may likely turn into parametric knowledge. RGB \cite{RGB} adopts a prompting approach by instructing: ``If the information in the document does not contain the answer, you will generate 'I cannot answer the question because of insufficient information in the documents.'" However, based on our experiments with TQA\cite{tqa} depicted in ~\ref{fig:comparison}, more powerful models with stronger adherence to commands are more likely to respond with "I cannot answer the question," making it challenging to intuitively assess the true retrieval-augmented generation (RAG) capability through prompting alone.

% \textcolor{red}{
To address the above-mentioned issues, we propose a multi-level fine-grained benchmark named Placeholder-RAG-Benchmark, emphasizing the following progressive dimensions: multi-level filtering, cross-entity and multi-attribute composition, and multi-paradigm reasoning. Besides, to decouple LLMs' parametric knowledge from the external knowledge, we formulate an innovative \textit{\textbf{Placeholder}}-based approach to decouple the contributions of the LLM's parametric knowledge and the external knowledge.

 %evaluating the LLM's capacity to filter noisy documents progressively; 
 %examining the LLM's proficiency in synthesizing information between multiple entities or attributes within single or multiple documents; 
%including contrastive and inductive reasoning based on both retrieved documents and the model’s parametric knowledge. 
% To address the above-mentioned issue, we propose a unified RAG evaluation benchmark to jointly quantify \textcolor{red}{retrieval accuracy, contextual integration fidelity, and generative reliability}. \textbf{Lack a short introduction of the proposed benchmark.}}
It is worth mentioning that, although our purpose is to evaluate the document utilization capability of the generative model in the RAG system, all the noise data associated with each piece of data can also be combined into a single library for evaluating retrieval models. This is because it involves many aspects (e.g., similar noise, multi-value, and multi-hop scenarios), which also present a significant challenge for retrieval models.

Our contributions are summarized as follows:

\textbf{(1) A More Granular Evaluation Framework.} We propose a novel Placeholder-RAG-Benchmark to enable fine-grained, multi-level assessments of LLMs within RAG systems. This framework introduces a more detailed dimensional structure for evaluating LLM capabilities, focusing on \textbf{(a) Multi-level Filtering}, \textbf{(b) Combination}, and \textbf{(c) Reference Reasoning} in effectively utilizing the retrieved corpus.

\textbf{(2) A Robust Placeholder Algorithm.} To achieve robust evaluations, a placeholder-based approach is formulated to mitigate the impact of the model's internal knowledge parameters, preventing LLMs from merely regurgitating answers and encouraging deriving conclusions from on RAG retrieved context.

\textbf{(3) Comprehensive RAG-based LLM Assessments.} Experiments demonstrate the limitations of representative LLMs in the RAG system's generation capabilities, particularly in error resilience and context faithfulness. Our benchmark provides a reproducible framework for developing more reliable and efficient RAG systems. The code and data will be released later.

\section{Related Work}
\textbf{RAG.} RAG is a framework that enhances the capabilities of large language models by integrating external knowledge retrieval with text generation. The foundational work by \cite{lewis2020retrieval} introduced RAG, combining a pre-trained retriever with a sequence-to-sequence generator, allowing models to condition outputs on retrieved documents. Subsequent studies expanded RAG’s applicability, such as REALM \cite{guu2020retrieval}, which incorporates retrieval into pre-training, and variants like RAG-Sequence \cite{shi2024generate} for multi-hop question-answering. Besides, some studies have begun to investigate multi-round retrieval augmentation. For instance, ICRALM \cite{ram2023context} and RETRO \cite{borgeaud2022improving} 
execute retrieval at the fixed token counts, while IRCot \cite{trivedi2022interleaving} triggers retrieval for each sentence.

\noindent\textbf{RAG Benchmark.} Recent advancements in RAG evaluation have produced diverse automated frameworks, yet standardization remains elusive. RAGAS \cite{es2024ragas} introduced LLM-based metrics (\textit{e.g.}, context relevance, faithfulness) via GPT-3.5 prompting, while ARES \cite{saad2023ares} proposed fine-tuned NLI models for similar assessments. Besides, \cite{chen2024benchmarking} developed robustness tests through context perturbation. \cite{adlakha2024evaluating} formulated heuristic algorithms for faithfulness estimation. RAGBench\cite{RGB} is one of the few evaluation frameworks for large model generation that addresses aspects such as noise resistance, integration, and counterfactual detection. However, these areas are evaluated with relatively coarse granularity, and the distribution of topics within the data is often skewed. 

% In incorporating these points, the paper aims to present a systematic and nuanced approach to evaluating the capabilities of RAG models, highlighting both the challenges and potential solutions in the field. This framework could serve as a foundation for future research, encouraging further refinement and adaptation to accommodate evolving model complexities.

%% file: sec/3_benchmark.tex
\section{Placeholder-RAG-Benchmark}
In this section, we will first introduce a unified benchmark to assess the information utilization abilities of  RAG-based LLMs from multiple perspectives. Then, we propose a dynamic placeholder substitution strategy to mitigate the influence of inherent knowledge encoded in pre-trained LLMs, ensuring that their performance solely relies on their ability to utilize external information. The overall pipeline shown in Figure \ref{fig:pls}.

\subsection{Triplet-based Meta-Data}
We obtain several types of popular entities from Wikipedia, such as Sports Events, Awards, Animals, Fictional Entities, and so on. From each type, a certain number of entities are selected as parent entities, denoted as $E^p$. To acquire semantically related entities, multiple propagation dimensions $D$ are generated for each parent entity, enabling the creation of child entities. Specifically, for a given parent entity $e^p$, the $i$-th propagation dimension $d_i$, it contains multiple child entities, with $j$-th child entity represented as $e^p_{ij}$. Using parent entities and their child entities within each propagation dimension, shared predicates are generated along with their corresponding values to form triplets $(e^p_{ij},p,v)$. After manual verification, the dataset consists of 224 parent entities, 2,272 child entities, and 16,033 triplets.

% This task evaluates the model's ability to identify accurate information.

\subsection{Fine-grained Evaluation Dimension}
For each dimension—\textbf{(a) Multi-level Filtering}, \textbf{(b) Combination}, and \textbf{(c) Reference Reasoning}—we define some basic evaluation tasks to quantitatively assess the fine-grained evaluation dimensions, where the evaluation data is constructed from the triplet-based data.

\subsubsection{Multi-Level Filtering.}
Multi-Level filtering measures the model’s ability to filter noise and extract accurate information.

For each child entity $e^p_{ij}$, its associated triplet $(e^p_{ij},p,v)$ is selected as the golden triplet and used to generate the corresponding query. Then, we construct varying levels of noise with different difficulty levels for fine-grained assessment. 

$\Delta$ \textit{Weak Noise: Noise from Completely Irrelevant Entities.} In this level, triplets from other parent entities are selected as weakly noisy triplets.

$\Delta$ \textit{Moderate Noise: Noise from Similar Entities.} Moderate noise triplets are generated by selecting multiple triplets $(e^p_{ij'},p,v)$, and their entities are from the same parent class and share a propagation dimension.

$\Delta$ \textit{Hard Noise: Noise from Generalizing Specific Situations.} Noise arising from general cases conflicting with specific scenarios. Triplets $(e^p,p,v)$ from the parent entity corresponding to the golden triplet are used as hard noise triplets.

\subsubsection{Combination.} 
% Cross-entity and multi-attribute composition aims to test the model's ability to retrieve information comprehensively (Information Exhaustiveness).

When dealing with multiple retrieved documents, it is critical to assess whether the model can gather comprehensive and relevant information for a given query. This ability is defined as combination abilities, a concept that entails detailed, cross-entity, and multi-attribute composition. We design diverse question types to thoroughly evaluate scenarios requiring extensive information aggregation. 

$\Delta$  \textit{Explicit Composition.} Two triplets $(e^p_{ij_1},p,v)$ and $(e^p_{ij_2},p,v)$ are selected from two child entities within the same diffusion dimension are selected as golden triplets. These are combined into a single question. 

\begin{framed}
\small
\noindent\textbf{Example of Explicit Composite Questions}\\
\noindent\textbf{Triplets:} (2016 Olympic, host country, Brazil) \& (2020 Olympic, host country, Tokyo)

\noindent\textbf{Question:} What are the host countries of the 2016 and 2020 Olympics, respectively?
\end{framed}

$\Delta$ \textit{Multi-Value Composition.} This level focuses on triples where the object contains multiple values, i.e., for a triplet $(e^p_{ij},p,v)$, $v$ takes the form $(v_1,v_2,...,v_n)$. The predicate is rephrased to denote an "include" relationship, resulting in triplets such as (2018 FIFA World Cup, host\_cities\_include, (Moscow, Sochi, Kazan, Saransk)). These are used to generate corresponding questions. In addition, values are then divided into smaller groups to form new triplets, e.g., $(e^p_{ij},p,(v_1,v_2))$ and $(e^p_{ij},p,(v_3,v_4))$, used as golden triplets.

$\Delta$  \textit{Multi-Scenario Composition.} When addressing broad or general situations, we aim for answers that consider multiple scenarios. To capture this, we generate a question using a parent entity triplet $(e^p,p,v)$. The golden triplets are derived from child entities within the same diffusion dimension that share the same predicates. This ensures that the model's response must incorporate information from multiple child entities.

% The comprehensiveness evaluation is designed based on golden triplets, while the noise robustness task is designed based on varying levels of noise triplets. These two evaluation frameworks are complementary rather than conflicting. For the comprehensiveness evaluation, noise is similarly introduced to the tasks using the aforementioned methods to ensure a complete and realistic assessment.

\subsubsection{Reference Reasoning.} 
Reference Reasoning evaluates the model’s ability to synthesize and construct reasoning chains using information when answers are not directly provided in the source documents (in contrast to the two previously mentioned abilities, which involve direct answering based on the documents). Reference Reasoning tasks are mainly multi-hop reasoning.

% \textbf{Reason-Pro Evaluation Task.} 
% This task evaluates the model’s ability to synthesize and reason using information when the source documents do not directly provide answers. The reasoning tasks are categorized into single-hop and multi-hop reasoning:

$\Delta$ \textit{Comparative Reasoning:} The model derives answers by comparing the values of specific attributes across two or more entities.

$\Delta$ \textit{Deductive Reasoning:}
This involves posing questions about attribute values through the use of a major premise and a minor premise. The process can be divided into two types:

\begin{framed}
\small
\noindent\textbf{Deductive Reasoning-1: Inheritance-based Attribute Questioning}\\
\noindent \textbf{Question:} How much does it cost to borrow books from Harvard?\\
\noindent \textbf{Document:} the book borrowing fees for Ivy League schools;\\
\noindent \textbf{Explanation:} Harvard belongs to the Ivy League schools, so the borrowing fees are subject to its parent.
\end{framed}

During construction, inheritance-based attribute questioning identifies subclass entities that inherit attributes from parent class entities. The constructed questions focus on querying the inherited attributes of the subclass entity, with the golden document providing an explanation of the corresponding attribute of the parent class entity.

\begin{framed}
\small
\noindent\textbf{Deductive Reasoning-2: Relationship-based Value Questioning}\\
\noindent \textbf{Question:} Is there a toll on the Guangzhou to Shenzhen expressway?\\
\noindent \textbf{Document:} Free expressways within the same provinces;\\
\noindent \textbf{Explanation:} The relation between Guangzhou and Shenzhen is the same province. So Free.
\end{framed}

During evaluation, relationship-based value questioning requires the following steps: (1) retrieving the relevant context according to the parent attribute of value, (2) find the relationship between entities via reasoning, and (3) obtaining the final answer via overall relationship-based reasoning. 

% The major premise is fixed, and the answer derives from the minor premise. This may involve identifying shared attributes across entities classified under different superclasses or subclasses.

$\Delta$ \textit{Comparative Deductive Reasoning:} This form of reasoning involves posing questions about the attribute sets of different entities. It requires a higher-level generalization as the major premise to guide the comparison. 
\begin{framed}
\small
\noindent\textbf{Comparative Deductive Reasoning}\\
\noindent \textbf{Question:} What safety measures do Tesla and Mercedes have?\\
\noindent \textbf{Document:} Tesla equipped with airbags and automatic alarms; Mercedes equipped with airbags and automatic alarms;\\
\noindent \textbf{Explanation:} Airbags and automatic alarms belong to safety measures, and they are common features
\end{framed}

\subsection{Dynamic Placeholder Substitution}
Given a query, dynamic placeholder substitution is to dynamically adjust the RAG contexts via changing the placeholder values. Then, we can observe whether LLMs can provide correct answers under these RAG contexts, thus effectively avoiding the internal knowledge of LLMs.

\textbf{Placeholder-based Context Generation.} We generate both the final golden documents and noise documents based on triplets. For each golden triplet, we replace its $v$ to create a placeholder-based triplet $(e,p,Placeholder)$, with multiple placeholders for multi-value triplets. Using GPT-4o and Qwen2.5-MAX, we then generate a corresponding document centered around the entity $e$, ensuring the information from $(e,p,Placeholder)$ is embedded within the document. Additionally, the document can be enriched by incorporating other predicates of the same entity, $(e,p',v)$. For noise documents, we directly synthesize corresponding documents for each noise triplet. Similarly, these documents can also be enriched using different predicates of the same entity.

\textbf{Placeholder-based Candidate Values Generation.} Based on $(e,p,Placeholder)$ and the question, we generate candidate values for the placeholder. These values must match the original value in the datatype. Furthermore, answers are constructed according to a set of placeholder values from the golden documents.

In fact, the dynamic placeholder substitution can enhance the robustness of the evaluation framework in two key ways: (a) By substituting critical information with placeholders, it reduces bias from the model’s internal knowledge during evaluation. (b) By repeatedly testing the same sample with minimal changes, it lowers the probability of models guessing correct answers.

% \subsection{Overview of Datasets.} 
\subsection{Benchmark Statistics} 
\textbf{Statistical Information.} We constructed an English dataset with 3,887 samples and a Chinese dataset with 3,387 samples. On average, the English dataset contains 1.74 golden documents per sample, while the Chinese dataset has 1.94. Regarding noise levels, the English dataset averages 5.80, 4.11, and 0.54 documents for weak noise, moderate noise, and hard noise, respectively. The corresponding values for the Chinese dataset are 1.94, 6.20, and 0.52. These ensure sufficient noise for robust experiments. Additionally, the English and Chinese datasets provide an average of 4.39 and 4.68 candidate placeholder values per sample, respectively, ensuring experimental stability. Further details on subtasks are provided in the Appendix \ref{sec:data_des}.

\textbf{Data Quality Validation.} Given the extensive use of GPT-4-based synthetic methods in our data construction pipeline, ensuring the quality of the generated data is of critical importance. Initially, we execute our evaluation pipeline using various state-of-the-art (SOTA) models under the simplest conditions—without introducing noise and with golden triplets provided as reference—to identify samples that multiple models fail to answer correctly (approximately 30\% of the dataset). Subsequently, these problematic samples undergo manual validation and correction. The manual validation process focuses on two primary objectives: (1) ensuring that substituting specific values for placeholders in the document does not introduce contextual inconsistencies, and (2) verifying the accuracy of the relationships between placeholders and their corresponding answers.

%% file: sec/4_exp.tex
\section{Experiment}

\begin{table*}
  \centering
  \small
  \caption{Performance of Various State-of-the-Art Models in this benchmark. Results unavailable in public reports are marked as "--". \textbf{Bold} values indicate the best experimental results among small-scale models, \textbf{\textit{italic bold}} values indicate the second-best experimental results, and \underline{underlined} values denote the third-best experimental results.}
  \label{tab:main}
  \begin{tabular}{l|cccc|cccc}
    \toprule
    Models & \multicolumn{4}{c|}{ZH} & \multicolumn{4}{c}{EN} \\
                            & \makecell{Overall} & \makecell{Multi-Level \\Filter} & \makecell{Combi-\\nation} & \makecell{Rea-\\soning} 
                            & \makecell{Overall} & \makecell{Multi-Level \\Filter} & \makecell{Combi-\\nation} & \makecell{Rea-\\soning}  \\
    \midrule
        Gemini-2.5-pro-preview    & \textbf{87.33}  & \textbf{97.92}  & \textbf{94.20}  & 70.18  & \textbf{84.89}  & \textbf{94.89}  & \textbf{85.32}  & \textbf{76.09}  \\
        Claude-3.7-sonnet         & \textit{\textbf{85.74}} & \textit{\textbf{97.62}} & \textit{\textbf{90.59}} & \textbf{70.39}  & \textit{\textbf{82.96}} & \textit{\textbf{93.18}} & \textit{\textbf{82.13}} & \textit{\textbf{76.51}}  \\
        Gemini-2.5-flash-preview  & \underline{81.85}  & \underline{93.92}  & \underline{88.54}  & 63.86  & \underline{79.20}  & \underline{90.69}  & \underline{80.30}  & 67.90  \\
        Qwen3-235B-A22B           & 80.76  & 94.92  & 88.18  & 60.23  & 78.68  & 90.56  & 78.32  & 69.97  \\
        Qwen3-30B-A3B             & 80.45  & 95.87  & 86.11  & 61.42  & 79.09  & 91.01  & 78.01  & 71.78  \\
        Deepseek-V3(241226)       & 77.54  & 94.58  & 81.00  & 60.32  & 79.02  & 89.91  & 77.18  & \underline{74.03}  \\
        Qwen3-235B-A22B w/o think & 75.20  & 91.50  & 79.67  & 57.14  & 70.27  & 83.95  & 66.37  & 67.15  \\
        Qwen-2.5-MAX              & 74.43  & 93.25  & 78.28  & 55.37  & 78.45  & 89.32  & 75.83  & 65.89  \\
        Qwen3-30B-A3B w/o think   & 71.05  & 91.08  & 72.22  & 54.76  & 65.38  & 84.76  & 61.12  & 58.47  \\
        Gemma3\_27b               & 70.24  & 92.21  & 73.09  & 50.24  & 79.18  & 92.03  & 78.00  & 71.33  \\
        Qwen3\_32B                & 69.69  & 89.75  & 75.74  & 46.70  & 78.05  & 90.69  & 77.23  & 69.65  \\
        Hunyuan-80B-A13B          & 68.84  & 93.50  & 68.94  & 50.64  & 73.42  & 86.89  & 71.58  & 66.38  \\
        GPT4.1                    & 66.26  & 89.75  & 71.95  & 41.27  & 60.79  & 84.76  & 64.02  & 35.37  \\
        Qwen2.5\_72B              & 64.87  & 92.92  & 64.99  & 44.14  & 68.90  & 87.01  & 64.30  & 63.69  \\
        GPT4o-1120                & 64.58  & 88.50  & 70.21  & 39.35  & 60.89  & 81.62  & 60.69  & 44.83  \\
        Gemma3\_12b               & 64.10  & 60.20  & 89.92  & 50.52  & 72.35  & 87.42  & 68.46  & 68.12  \\
        Qwen3\_8B                 & 63.04  & 86.87  & 67.49  & 39.47  & 76.80  & 88.36  & 76.27  & 68.71  \\
        Qwen3\_32B w/o think      & 60.73  & 89.50  & 59.53  & 41.30  & 68.30  & 84.35  & 63.74  & 64.59  \\
        Qwen2.5\_32B              & 58.76  & 92.00  & 51.33  & 44.60  & 66.70  & 85.66  & 63.04  & 58.92  \\
        Qwen2.5\_14B              & 55.94  & 89.42  & 52.69  & 35.87  & 63.29  & 84.40  & 57.35  & 58.34  \\
        Qwen2.5\_7B               & 49.31  & 83.29  & 47.47  & 26.92  & 63.16  & 81.90  & 56.76  & 61.00  \\
        Qwen3\_8B w/o think       & 50.02  & 83.96  &  47.83 & 28.17  & 64.71  & 83.21  & 58.93  & 61.52  \\
        Gemma3\_4b                & 47.67  & 78.33  & 37.41  & 39.26  & 57.58  & 77.98  & 48.50  & 59.41  \\
    \midrule
  \end{tabular}
\label{tab:main}
\end{table*}

In this section, we will conduct some relevant experiments to demonstrate the effectiveness of different proficiency tests and the performance of different models under the hyperparameter configurations we recommend. 

Through the evaluation of various large language models (LLMs) using our benchmark, we draw the following conclusions:

1. \textbf{Impact of reasoning modes:} Models exhibit significant performance differences across our three evaluation dimensions when operating in "reasoning" versus "non-reasoning" modes. Analysis of reasoning chains indicates that reasoning models effectively process and reflect on noisy document hierarchies, thus aggregating all relevant information in combination tasks and performing complex reasoning in reference-reasoning tasks, while non-reasoning models are prone to selecting highly misleading answers without critical analysis.

2. \textbf{Filter ability vs. model size:} Filter capability does not scale linearly with model size. Smaller models sometimes outperform larger ones in "needle-in-a-haystack" tasks by directly extracting verbatim text from source documents, thereby matching key answer phrases more precisely. In contrast, larger models often paraphrase source text, occasionally omitting critical details. For instance, GPT-4 models tend to simplify date information, reporting only "month-year" while omitting "day," despite the day being explicitly provided in the source.

3. \textbf{Performance on combination and inference tasks:} Larger models demonstrate clear advantages in combination and reference inference tasks, excelling at aggregating, reasoning, and constructing complete reasoning chains. Among the tested models, the Gemma series showed stronger reasoning capabilities, while the Qwen series exhibited superior performance in information integration and combination tasks.
% The overall results are depicted in Table \ref{tab:results_all} and some bad cases are shown in Figure \ref{fig:bad_case}. 

% \begin{figure}[t]
%   \includegraphics[width=\columnwidth]{fig/bad case.png}
%   \caption{Error Cases of Three Dimension by GPT-4o-1120.}
%   \label{fig:bad_case}
% \end{figure}

\begin{table*}[h]
\centering
\small
\caption{Error Cases of Three Dimension by GPT-4o-1120.}
\begin{tabular}{|m{2cm}|m{5cm}|m{5cm}|m{5cm}|}
\hline
& \textbf{Multi-Level Filtering} & \textbf{Combination} & \textbf{Reference Reasoning} \\
\hline
\textbf{Question} & When did Mozart perform the Turkish March? & What is the shape of eucalyptus leaves? & Which country can the Best Mentor-Disciple Duo in East Asian martial arts visit? \\
\hline
\textbf{Answer} & \textbf{November 9, 1786} & \textbf{Lanceolate \& Oval} & \textbf{the United States} \\
\hline
\textbf{Documents} & 
\parbox{4.5cm}{\textit{Golden Document} \\ Mozart's performance of the Turkish...it was brilliantly presented at the Vienna Conservatory on \textcolor{blue}{\textbf{November 9, 1786}} \\ \textit{Noisy Document} \\ The Turkish March,... When it was performed by Bobby McFerrin... The date of this performance was \textcolor{red}{\textbf{October 7, 1783}}} & 
\parbox{4.5cm}{\textit{Golden Document}\\...In addition, the blue gum tree has unique leaf shapes, including \textcolor{blue}{\textbf{lanceolate}} ones. \\ \\ ...The leaves of eucalyptus trees in the Northwest region are usually \textcolor{blue}{\textbf{oval-shaped}}} & 
\parbox{4.5cm}{\textit{Golden Document} \\ The Best Mentor-Disciple Duo in Karate can travel to \textcolor{blue}{\textbf{the United States}}. \\ \textit{Noisy Document} \\ The Best Mentor-Disciple Duo in Wrestling can visit the Shaolin Temple in \textcolor{red}{\textbf{China}} for a pilgrimage.} \\
\hline
\textbf{Response} & \textcolor{red}{\textbf{October 7, 1783}} & The leaves of the blue gum tree are mainly \textcolor{blue}{\textbf{lanceolate}} & \textcolor{red}{\textbf{China}} \\
\hline
\end{tabular}
\end{table*}

\subsection{Experimental Setup}
\textbf{Hyper-parameters.} We employed a configuration where both l1 noise doc and l2 noise doc were set to 4, and l3 noise doc was set to 1. Three placeholders were used for the evaluation in our main experiments. All models were tested on both Chinese and English datasets. For closed-source models, we accessed them via APIs, while for open-source models, experiments were conducted using 8 NVIDIA A100 GPUs with a batch size of 16. The VLLM framework was used to implement and run the open-source models.

\textbf{Evaluated LLMs.} We utilized several state-of-the-art models of varying sizes and architectures, including GPT-4o, Claude 3.7, Gemini, Qwen Series and DeepSeek v3. Notably, since the Qwen3 series incorporates an automatic reasoning mechanism, we evaluated the scenarios under two modes—enabled and disabled reasoning—for the these models.

\begin{algorithm}[!t]
\caption{Evaluation Pipeline}
\label{algo-one}
\begin{algorithmic}[1]
    \STATE \textbf{Input:} LLM, Placeholder-RAG-Eval Dataset $D$, $n<N$ as candidate placeholder number
    \FOR{each data point $d \in D$}
        \FOR{$i$ in range($n$)}
            \STATE result$_i$=infer(LLM, $d$\&$P_i$)
        \ENDFOR
        \STATE Score=AVG(metric(result$_{\{1,\dots, N\}}$,GT))
    \ENDFOR
     \STATE Calculate the final evaluation score, aggregated by different RAG tasks $T_i$
    \RETURN Scores$_{T_i}$
\end{algorithmic}
\end{algorithm}
\textbf{Evaluation Pipeline.} 
Let $D, P_N$ denote one data point(\textit{i.e.}, query, RAG context, answer) and the placeholder set, respectively, where $N$ denotes the number of candidate placeholder values.
The overall evaluation pipeline is illustrated in Algorithm \ref{algo-one}. During evaluation, hyperparameters are set to define the number of placeholders per data point and the configuration of noise documents. The algorithm iteratively samples data points by replacing placeholders, computes inference results for each placeholder, and calculates an average score per sample. Finally, it aggregates evaluation scores across different RAG tasks.

\textbf{Evaluation Metrics.} For evaluation, we mainly use two metrics: accuracy (Covered Exact Match) and GPT evaluation.

Accuracy refers to whether the keywords appear in the document. We use logical operators "or ($|$)" and "and (\&)" to improve the accuracy rate. The scenarios for using the "and" condition include situations where multiple possible answers must all be mentioned, and when a relatively long phrase is broken down into key, small elements.

As for GPT evaluation, we directly let GPT4o judge whether the answer is correct according to the question.

\subsubsection{Placeholders}
This part primarily examines the role of placeholders in evaluation methods. Experiments were conducted across different scenarios using three placeholders to test various models. The study explored the performance differences of multiple placeholders as well as the stability of the answers. To facilitate the tests, we utilized models of varying sizes from the Qwen2.5 series to observe the changes in outcomes when three placeholders were modified for the same dataset. Specifically, these outcomes included scenarios where all answers were correct, all answers were incorrect, and some answers were incorrect. The results reveal that as model size increases, the proportion of completely correct answers rises, while the instances of partial errors and complete errors decrease; notably, the proportion of partial errors declines significantly slower than the proportion of complete errors.

\begin{figure}[t]
  \includegraphics[width=\columnwidth]{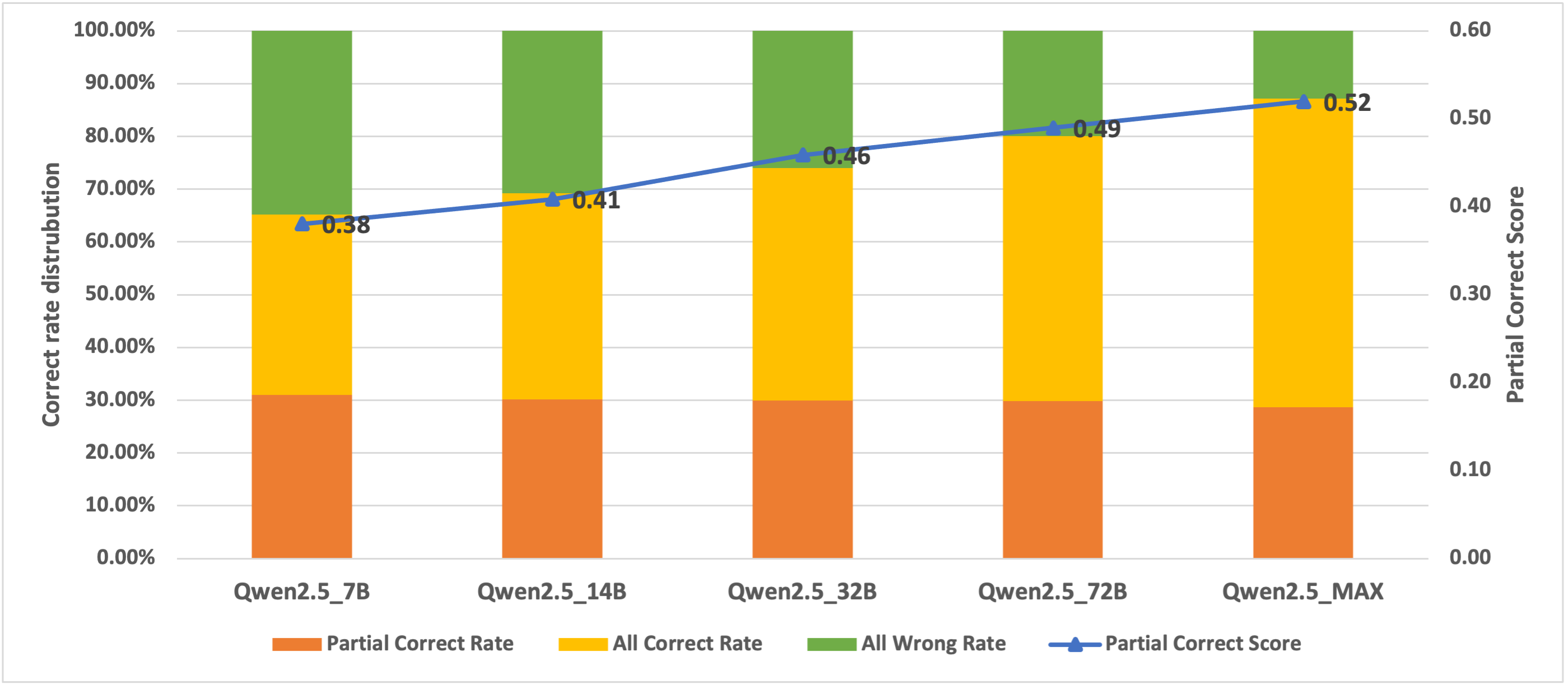}
  \caption{Qwen2.5 Series Model Performance Comparison, among these, the line graph for "partial correct score" represents the specific scores achieved in partially correct cases.}
  \label{fig:placeholders}
\end{figure}

Furthermore, analyzing placeholder-level performance, particularly for partially correct outcomes, reveals that the proportion of such answers remains relatively stable at around 30\%. This indicates that placeholders influence models of all sizes. However, as shown in Figure \ref{fig:placeholders}, the proportion of partially correct answers increases steadily with model size, suggesting that larger models exhibit greater stability when handling variable RAG documents.

\subsection{Experiment About Assessment Dimension}
In this section, we examine the effectiveness of the proposed evaluation dimensions. To ensure better control of variables, we conducted experiments using Qwen-2.5 models of varying sizes to analyze how performance varies across the three defined dimensions: \textbf{(a) Multi-level Filtering}, \textbf{(b) Combination}, and \textbf{(c) Reference Reasoning}, as reflected in our evaluation dataset.

\begin{table}[h]
\centering
\small
{
\begin{tabular}{lcccc}
\toprule
\textbf{Noise Config} & Config 1& Config 2 &Config 3 & Config 4 \\
\textbf{Filtering-en} &  &   &  &  \\
\midrule
\texttt{Qwen2.5 7B}  & 89.01 & 83.74 & 85.09 & 84.72  \\
\texttt{Qwen2.5 32B} & 82.68 & 80.80 & 80.35 & 79.34  \\
\texttt{Qwen2.5 72B} & 89.09 & 88.15 & 87.58 & 87.42  \\
\texttt{Qwen2.5 MAX} & 85.38 & 84.84 & 84.68 & 84.56  \\
\bottomrule
\end{tabular}
}
\caption{The model's performance under different noise configurations is evaluated on the filter task. Specifically, Config 1 represents the noise ratio of Weak:Moderate:Hard as 5:0:0. Config 2, Config 3, and Config 4 correspond to ratios of 3:2:0, 1:4:0, and 1:3:1, respectively.}
\label{tab:noise_config_en}
\end{table}

\begin{table}[h]
\centering
\small
{
\begin{tabular}{lcccc}
\toprule
\textbf{Noise Config} & Config 1& Config 2 &Config 3 & Config 4 \\
\textbf{Filtering-zh} &  &   &  &   \\
\midrule
\texttt{Qwen2.5 7B}  & 95.33 & 84.54 & 83.92 & 80.83  \\
\texttt{Qwen2.5 14B} & 95.28 & 85.62 & 84.38 & 80.96  \\
\texttt{Qwen2.5 32B} & 95.64 & 88.93 & 88.24 & 85.13  \\
\texttt{Qwen2.5 72B} & 95.58 & 93.71 & 94.17 & 92.92  \\
\texttt{Qwen2.5 MAX} & 95.87 & 94.17 & 93.33 & 93.13  \\
\bottomrule
\end{tabular}
}
\caption{Results on different noise ratios on the Chinese dataset.}
\label{tab:noise_config_zh}
\end{table}

\begin{table}[h]
\centering
\small
{
\begin{tabular}{l|c|ccc}
\toprule
{\textbf{Model}} & Overall &\multicolumn{3}{c}{Combination}\\ \cline{2-5}
 &(\%)& L1 (\%) & L2 (\%) & L3 (\%)\\ 
\midrule
\texttt{Qwen2.5 7B}  & 47.47 & 74.07 & 39.66 & 28.60 \\
\texttt{Qwen2.5 14B} & 52.69 & 79.27 & 47.00 & 31.73 \\
\texttt{Qwen2.5 32B} & 51.33 & 85.07 & 45.32 & 23.53 \\
\texttt{Qwen2.5 72B} & 64.99 & 84.00 & 56.36 & 54.53 \\
\texttt{Qwen2.5 MAX} & 78.28 & 86.53 & 73.06 & 75.20 \\
\bottomrule
\end{tabular}
}

\caption{The performance of Qwen2.5 series models on combination tasks across various english datasets: L1 denotes explicit composite questions, L2 refers to multi-value questions, and L3 represents multi-scenario analysis Questions.}
\label{tab:model_Composition_scores}
\end{table}

In the context of \textbf {Multi-Level Filtering} tasks, we assess the influence of noise at varying levels of difficulty on model performance. To ensure controlled experimentation, we systematically adjust the proportions of noisy documents in the filter task. From Table \ref{tab:noise_config_en} and Table \ref{tab:noise_config_zh}, we observe that increasing moderate and hard noise levels degrade model performance on even simple filter tasks, while keeping the total number of noisy documents constant. This demonstrates the effectiveness of our multi-level noise design. Moreover, we observe that while the 7B model performs well in scenarios with only weak noise, its performance deteriorates significantly as noise becomes more challenging. In contrast, larger models demonstrate greater resilience to handling more complex noise.

% \begin{table}[h]
% \centering
% \resizebox{\linewidth}{!}{
% \begin{tabular}{lcccc}
% \toprule
% \textbf{Model} & \makecell{Overall \\ (\%)} & \makecell{Composition\\ L1 (\%)} & \makecell{Composition\\ L2 (\%)} & \makecell{Composition\\L3 (\%)} \\
% \midrule
% \texttt{Qwen2.5 7B}  & 47.47 & 74.07 & 39.66 & 28.60 \\
% \texttt{Qwen2.5 14B} & 52.69 & 79.27 & 47.00 & 31.73 \\
% \texttt{Qwen2.5 32B} & 51.33 & 85.07 & 45.32 & 23.53 \\
% \texttt{Qwen2.5 72B} & 64.99 & 84.00 & 56.36 & 54.53 \\
% \texttt{Qwen2.5 MAX} & 78.28 & 86.53 & 73.06 & 75.20 \\
% \bottomrule
% \end{tabular}
% }
% \caption{The performance of the Qwen2.5 series models on Composition tasks across various English datasets: Composition\_L1 denotes explicit Composite Questions, Composition\_L2 refers to Multi-Value Questions, and Composition\_L3 represents Multi-Scenario Analysis Questions.}
% \label{tab:model_Composition_scores}
% \end{table}

In the \textbf{Combination} task, we experiment with three distinct task types. As shown in Table \ref{tab:model_Composition_scores}, for L1 tasks, smaller models like Qwen2.5-7B perform well, whereas for the second and third task types, larger models exhibit a significant advantage.

The results about \textbf{Reference Reasoning} are shown in Table \ref{tab:model_rag_scores}. There is no absolute measure of difficulty among its four aspects. However, we can observe that larger models generally possess stronger overall document reasoning capabilities. Nevertheless, on the whole, the category of deductive reasoning is more challenging compared to other scenarios. On the one hand, in deductive reasoning, the model needs to identify the required attributes and then find the common attributes. It requires numerous cognitive leaps to arrive at the correct answer, posing a significant challenge to the model.

\begin{table}[ht]
\centering
\large
\resizebox{\linewidth}{!}{
\begin{tabular}{lccccc}
\toprule
\textbf{Model} & \makecell{Overall \\Reasoning (\%)} & \makecell{Comp\\ (\%)} & \makecell{Comp\\Deduc(\%)} & \makecell{Deduc\\V1 (\%)} & \makecell{Deduc\\V2 (\%)} \\
\midrule
\texttt{Qwen2.5 7B}  & 26.92 & 32.97 & 20.31 & 26.11 & 41.83 \\
\texttt{Qwen2.5 14B} & 35.87 & 42.65 & 30.79 & 33.67 & 48.67 \\
\texttt{Qwen2.5 32B} & 44.60 & 53.05 & 36.81 & 43.33 & 62.00 \\
\texttt{Qwen2.5 72B} & 44.14 & 48.39 & 33.67 & 45.33 & 66.50 \\
\texttt{Qwen2.5 MAX}        & 55.37 & 67.38 & 46.16 & 55.11 & 73.17 \\
\bottomrule
\end{tabular}
}
\caption{The performance of the Qwen2.5 Series Models on Multi-Paradigm Reasoning: Comp denotes Comparative Reasoning, Comp Deuc refers to Comparative Deductive Reasoning, Deduc v1 represents Inheritance-based Attribute Deductive Reasoning, and Deduc v2 indicates Relationship Based Deductive Reasoning.}
\label{tab:model_rag_scores}
\end{table}

%% file: sec/5_conclusion.tex
\section{Conclusion}
This paper introduces PRGB, a placeholder-based RAG benchmark designed to evaluate the information utilization capabilities of from multiple dimensions. Starting from triples, we constructed two human-reviewed evaluation datasets: 3,887 English samples and 3,387 Chinese samples, synthesized using GPT-4. The benchmark comprehensively assesses models across three dimensions: multi-level filtering capability, complex composition capability, and multi-paradigm reasoning capability. To enhance evaluation robustness, we employ a dynamic placeholder strategy that replaces critical information in reference documents during testing, mitigating the interference of knowledge stored in the model's internal parameters. In the experiments, we conducted a thorough evaluation of various state-of-the-art (SOTA) models, offering insights into model selection for RAG scenarios. Furthermore, ablation studies on the Qwen 2.5 series models across different tasks verified the effectiveness of the multi-level noise mechanism and the placeholder strategy. Looking ahead, we aim to optimize evaluation metrics and establish a more comprehensive evaluation framework to better benchmark RAG-based generative tasks.